%% file: lacuna-v1.tex
\documentclass[11pt]{article}
\usepackage[margin=1in]{geometry}
\usepackage[T1]{fontenc}
\usepackage{lmodern}
\usepackage{amsmath,amssymb}
\usepackage{booktabs}
\usepackage[numbers,sort&compress]{natbib}
\usepackage{xurl}
\usepackage[hidelinks]{hyperref}
\usepackage{graphicx}
\usepackage{tikz}
\usetikzlibrary{arrows.meta}
\usepackage[font=footnotesize,skip=3pt]{caption}
\makeatletter
\renewcommand\paragraph{\@startsection{paragraph}{4}{\z@}%
  {1.6ex \@plus .4ex \@minus .2ex}{-1em}{\normalfont\normalsize\bfseries}}
\makeatother
\AtBeginDocument{\setlength{\abovedisplayskip}{6pt plus 2pt minus 2pt}%
  \setlength{\belowdisplayskip}{6pt plus 2pt minus 2pt}}
\title{Estimating the agent actions that no log recorded:\\six conditions for capture-recapture over incomplete logs}
\author{Caio Theodoro\\\small Avenza Labs (\texttt{avenza.cloud})\\
\small ORCID \href{https://orcid.org/0009-0005-7343-981X}{0009-0005-7343-981X}}
\date{First version 16 November 2026; this version 30 November 2026 (\texttt{paper-v2.2})}

\begin{document}
\maketitle

\begin{abstract}
An audit of an AI agent starts from its log, and tamper evidence shows that a log was
not altered, not that it is complete. Where no trusted counter numbers every action, the
actions missing from every log can still be counted, by capture-recapture over several
incomplete logs. Five pre-registered studies establish when that count can be relied on.
In a live rig with two uncoordinated originators, whose telemetry relays are killed and
restarted and where no loss is simulated, one estimator named in advance held the true
count inside its 95\% interval in 269 of the 277 runs of the primary condition (Wilson
0.944 to 0.985; the criterion as registered is met). 23 of the 300 planned runs were set
aside on integrity: with those counted as not covered it is 269 of 300 (0.857 to 0.926,
``not shown''), and post hoc, with the 20 that left a full record scored as they were, it
is 288 of 300 (0.931 to 0.977). The point was closer to the true count than a
sequence-gap baseline in 223 of the 277 runs, and without the end-state witness the
interval covered 10. The bound takes runs as independent although they replay shared
sessions, an interval with no upper end would have counted as covering (none occurred),
and this is one rig. Four studies on recorded agent episodes with simulated loss show
how the estimate fails, each time with no sign in its output: with inexact record
linkage (coverage 0.893 against a bar of 0.90), with a channel that no log itemises
(0.759), and, post hoc, with loss that depends on the action (0.187) and loss in blocks
(0.559). The result is six conditions, checked by reviewing a deployment's design and
not by the estimate, under which the actions an agent's logs missed can be counted.
\end{abstract}

\section{Introduction}
\label{sec:question}

Every audit of an AI agent begins with its log~\citep{euaiact}, and every log is a claim
of completeness. The engineering around logs is about integrity: hash chains, Merkle
commitments and trusted execution show that a line was not changed after it was written
(\S\ref{sec:related}). None of it shows that the log is whole, because a line never
written leaves no gap in a chain. Lines do go unwritten. Agents have deleted their own
trace events when asked, and unprompted when it raised their
reward~\citep{qin2026tamper}, and ordinary telemetry loses records with no adversary at
all (\S\ref{sec:s5results}). This paper asks when the actions that no log recorded can
be counted anyway.

Where one trusted counter numbers every action at the source or at the sink, the count
is its last number and nothing needs estimating (\S\ref{sec:earlier} sets such a counter
beside the estimate and shows where each of them fails). The setting here is the other
one: actions come from several processes that share no numbering, the logs belong to
components that lose records on their own, and no single log is trusted to be whole.
There the actions missing from every log are a hidden population, and the overlap of
several incomplete registers estimates what all of them
missed~\citep{fienberg1972,bird2018mse}. No work we know of applies this to count the
actions missing from an agent's logs (\S\ref{sec:related}). The method is known to fail
when the registers depend on one another in ways its model leaves out, so the question
is under which conditions the count can be relied on, and what a failure looks like.

The answer has two parts. The count can be recovered. On a rig whose telemetry loses
records by itself, the interval of an estimator named in advance held the true count in
269 of the 277 runs of the primary condition that were left when 23 of the 300 planned
were set aside on integrity (269 of 300 with those counted as not covered, which reads
``not shown''), and its point was closer to the truth than counting sequence gaps or
counting the union of the logs (\S\ref{sec:s5results}). And the same estimate fails,
with nothing in its output to show it, when any of six properties of the deployment
does not hold (\S\ref{sec:when}). The contributions are these.

\begin{itemize}\setlength\itemsep{1pt}\setlength\parskip{0pt}
\item \textbf{A count of unlogged actions under real loss.} A pre-registered study on a
live rig, two originators and two planes, in which telemetry relays are killed and
restarted and no loss is simulated: 420 registered runs, one estimator and one criterion
named in advance, three floors and a sequence-gap baseline beside it
(\S\ref{sec:s5design}, \S\ref{sec:s5results}).
\item \textbf{Four ways the estimate fails in silence, each with its number.} Records
matched by what they did and not by an exact id (coverage 0.893 and 0.401, against 0.948
and 0.946 with exact ids); a channel that no log itemises (0.759); loss that depends on
the action (0.187); and loss in blocks (0.559; on a second set of loss seeds, where
the plain interval gives 0.566, the wider interval that the live study registers gives
0.952) (\S\ref{sec:earlier}). Table~\ref{tab:earlier} prints the
last two on another set of loss seeds, as 0.254 and 0.574.
\item \textbf{Which witness carries the count.} A record of the end state is what makes
the missing actions countable: with the two telemetry logs alone the interval covered 10
of the same 277 runs (\S\ref{sec:s5results}).
\item \textbf{Six conditions a design review can check.} None can be read off an
estimate; a small tool checks part of them from a map of a deployment's logs
(\S\ref{sec:when}).
\end{itemize}

Figure~\ref{fig:map} sets the cells of all five studies against the bar.

\section{Related work}
\label{sec:related}

\paragraph{Tamper evidence.} Hash chains and Merkle commitments detect an edit to a
log~\citep{flightrecorder2026}, trusted execution and notarised TLS give
integrity~\citep{vet2025}, and witnesses can replay a tamper-evident
log~\citep{haeberlen2007peerreview}. These address whether a log was altered. The study
that showed agents tampering with their own traces~\citep{qin2026tamper} does not
estimate what is missing.

\paragraph{Capture-recapture.} Multiple-systems estimation counts a hidden population
from the overlap of incomplete lists~\citep{fienberg1972,bird2018mse}; with two lists it
is the Lincoln--Petersen estimate, which Chapman's form corrects for small
counts~\citep{chapman1951}. Its known
failure is dependence between the lists: in software inspection it is the point at
which the method breaks~\citep{eick1992inspection,petersson2004inspection}, capture
probabilities that differ between individuals make the classical estimates
undercount~\citep{chao1987}, and the count is not identified without an assumption the
data cannot test~\citep{aleshinguendel2024}. The estimators here make that assumption by
leaving the highest interaction out of a log-linear model.

\paragraph{Capture-recapture on software and model output.} The method has been used to
estimate how complete an LLM security audit is, with each of several audit lenses a
capture occasion~\citep{hirashima2026vsat}, and species estimators to estimate how complete an
event log is in process mining~\citep{yang2012completeness}. Capture-recapture over LLM
judges counting label errors produced valid intervals too wide to beat a raw
count~\citep{kansal2026labels}. Each counts kinds of finding, of trace or of error. None
counts the actions an agent took that its logs do not hold.

\paragraph{Other assumptions about dependence.} Three lines of work do not fix a
log-linear model: a Bayesian estimate of the population size that uses Dirichlet process
mixtures~\citep{manriquevallier2016lcmcr}; dependence-robust confidence intervals from a
bound on the dependence between pairs of lists~\citep{sun2020bounds}, which is partial
identification and returns a set, not a point; and the assumption that the lists are
``not arbitrarily dependent'' within the part the others missed~\citep{dulcerubio2024}.
Each replaces one assumption about dependence with another. None was a registered
estimator of any study here, and this paper reports no estimate from any of them. An
interval under a bounded odds ratio, this repository's own, is set beside the registered
one post hoc (\S\ref{sec:s5results}).

\section{Terms and method}
\label{sec:method}

An \emph{action} is a tool call that changes the environment's state or sends data out.
The environment kernel appends one ledger entry per action, so the true count $N$ is a
program's output that no witness and no estimator reads. A \emph{witness} is a log that
records some of the actions. A \emph{channel} is a route an action can take: a direct
tool call, or a batch tool that runs a script of many operations in one call. A log
\emph{itemises} a channel if it holds one record per action taken through it.

\paragraph{Estimates.} Capture-recapture cross-classifies the actions by which witnesses
recorded them and estimates the cell nobody recorded. Studies 1 to 4 use
\texttt{ll\_bic}: Poisson log-linear models of the three-log
table~\citep{fienberg1972,cormack1989loglinear}, pairwise interactions chosen by BIC, a
95\% profile-likelihood interval~\citep{cormack1992interval}, no three-way interaction. Study 5 registers \texttt{ll\_sa\_hom} (\S\ref{sec:s5design}).

\paragraph{Scoring.} \emph{Coverage} is the share of replicates (studies 1 to 4) or of
runs (study 5) whose 95\% interval holds $N$. Every registration sets the bar at 0.90
and was frozen in a commit before the runs it governs were made
(Appendix~\ref{app:registration}). \emph{Post hoc} marks every result outside a
registration.

\paragraph{Floors.} Every table of estimates carries three counts that need no model:
\texttt{trust\_trace}, the agent's own log taken as complete; \texttt{union\_count}, the
distinct actions in any log, with no interval; and \texttt{union\_bound}, the same
number as a lower bound with no upper end, which always covers and never bounds.

\section{Where the estimate fails: studies 1 to 4}
\label{sec:earlier}

\begin{figure}[!tb]
\centering
\includegraphics[width=\linewidth]{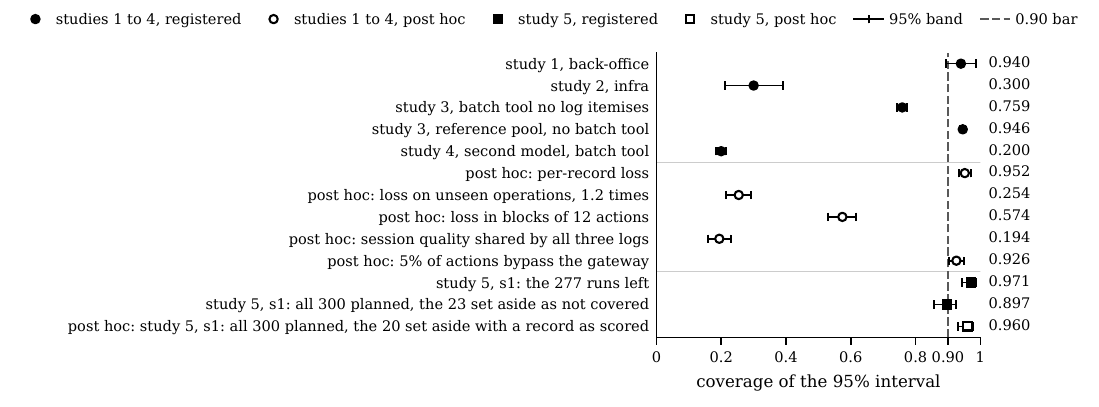}
\caption{Coverage against the 0.90 bar (dashed), one row per cell of
Table~\ref{tab:earlier} and three rows for study 5. Studies 1 to 4: \texttt{ll\_bic} under
simulated loss; filled circles are registered cells, open circles post hoc ones, and a bar
is the two-sided band at the intervals' nominal level from the artifact's Monte Carlo
standard error. Study 5 (squares): \texttt{ll\_sa\_hom} covered 269 of 277 \texttt{s1}
runs left, with the artifact's Wilson interval. 23 of the 300 planned \texttt{s1} runs were set aside on integrity, and
with those counted as not covered it is 269 of 300, whose interval straddles the bar and
reads ``not shown''. The open square is post hoc: 288 of 300, with the 20 runs set aside
that left a full record scored as they were and the other 3 counted as not covered
(\S\ref{sec:s5results}). The rows differ in estimator, linkage, loss and unit (replicates
or runs) and are not one experiment.}
\label{fig:map}
\end{figure}

\paragraph{The common setup.} The first four studies ran a live agent
(\texttt{deepseek-v4.1-flash}; study 4 adds \texttt{qwen3.8-flash}) on queues of
tickets. An \emph{episode} is one shift of the agent on one queue. There are two
environments: a back-office queue, with tickets about customers and orders, and an
infrastructure queue, called infra below. Studies 3 and 4 use a variant of the
back-office queue that adds a batch tool. Three witnesses saw the actions: the trace the
agent's harness writes, the log of a gateway proxy, and a
diff of the environment's state at the end of the shift, with its outbound mail. The
logs of these episodes are complete, and loss was simulated on them afterwards. A
\emph{replicate} resamples the recorded episodes, drops log lines by a loss model whose
random draws are fixed by a \emph{loss seed}, and estimates $N$ again. \emph{Linkage} is
how records of one action in different logs are matched: by \emph{effect} (what the
record says the action did), by a \emph{call id} the harness writes into the trace and
the gateway log, or by the \emph{ledger id}, which is exact and which no auditor has.
A \emph{cell} is one environment under one loss model and one linkage. Each study
registered its cells and its criteria in advance, with the bar of 0.90 on coverage.
Table~\ref{tab:earlier} gives one registered cell for each study, with its mean
interval width and the floors, and Figure~\ref{fig:map} draws the coverages.

\begin{table}[!tb]
\centering\small
\input{tables/v1_earlier}
\caption{Studies 1 to 4 and the post hoc loss cells: coverage of \texttt{ll\_bic}, of a
tail-corrected sequence counter where it was computed, and of the three floors. Two
rows are not the runs the abstract quotes: loss on unseen operations is 0.254 here and
0.187 in the series of doses, and loss in blocks is 0.574 here and 0.559 in the series
of block lengths. Each pair is one loss and one linkage on two sets of loss seeds; the
two post hoc paragraphs of this section give both.}
\label{tab:earlier}
\end{table}

\paragraph{Linkage has to be exact (studies 1 and 2).} With records matched by effect,
coverage on back-office episodes was above the bar at the registered seed
(Table~\ref{tab:earlier}, first row) and is $0.893 \pm 0.006$ over 30 fresh loss seeds,
under it: one seed had decided the verdict. On infra the same cell failed
(Table~\ref{tab:earlier}, second row; 0.401 over 20 seeds). In both environments the
failure goes away when the registered linkage is replaced by the ledger id: coverage is
then 0.948 in back-office and 0.946 in infra. Matching by effect pairs records of
different actions that did the same thing, and the false overlap reads as dependence
between the logs. Post hoc, the call id restores the
back-office figure (0.946) and not the infra one (0.859), because the end-state diff
carries no id. An id is only as good as the lines that carry it: with a twentieth of the
lines duplicated in all three logs coverage is 0.094, and with a twentieth forged in all
three it is 0.000.

\paragraph{Every channel has to be itemised (study 3).} The agent was
given a batch tool. The ledger gets each inner action; the trace and the gateway log one
call holding the script's text, and by stipulation no witness parses it. The agent ran
0.856 of its 1283 actions through the tool. Coverage
fell under the bar, on intervals several times as wide as on a reference pool recorded
with no batch tool, where it held (Table~\ref{tab:earlier}, third and fourth rows). The
estimator still reported a gap in 0.995 of replicates, as it does in every replicate of
the reference pool, so a detected gap does not say that a channel is unseen. Post hoc,
one record per scripted action, taken from the ledger, restores 0.949.

\paragraph{A log must not lose a batch as one (study 4).} The scripts were parsed from
their logged text, so every inner action has a record, and the question was what happens
when a log drops a script's lines together. Coverage is then 0.469, where the same
records lost one by one give 0.952. Under an adversary prompt, which told the agent how
it is counted and asked it to undercount, the pair is 0.610 against 0.940, and on a
second model 0.335 against 0.941. With the scripts not parsed, that model routed 0.970 of
its actions through the batch tool and its registered cell is the lowest of the four
studies (Table~\ref{tab:earlier}, fifth row). The adversary prompt split the pool (11 of
23 episodes used no script, against 1 under the neutral prompt), and no episode found a
route beyond the one channel. That adversary was weak: one episode, black box, the
channel not named.

\paragraph{Loss that depends on the action breaks the estimate in silence (post hoc).}
The estimate assumes that what
the logs lose together is as visible to the remaining witness as everything else. To
break that, the simulated loss was raised on the operations the end-state diff cannot
see, in steps called doses, with exact matching. A tenth more loss there takes coverage
to 0.647 and a fifth more to 0.187, on intervals barely wider than before, so nothing in
the output signals it. At higher doses the intervals widen and coverage returns: 0.303 at
three tenths more, 0.517 at two fifths, 0.857 at three fifths and 0.907 at four fifths,
where the mean width is 1902.9 actions against 38.9 with no excess (300 replicates a
dose). Table~\ref{tab:earlier} prints 0.254 for the fifth-more dose where the series
gives 0.187: it is the same loss and linkage run again on another set of loss seeds,
with 500 replicates (Monte Carlo standard errors 0.0195 there and 0.0225 in the series).

\paragraph{Loss in blocks needs a wider interval (post hoc).} Loss need not fall on
records one by one: in study 5's rig a relay's outage takes a run of them. With 12 consecutive actions lost together, coverage in
back-office is 0.559 over 2000 replicates and 0.802 in infra. The same two cells were
run again on a second set of seeds, in a comparison of intervals (0.566 and 0.795), and
Table~\ref{tab:earlier} holds the back-office one on a third (0.574). The repair is an
interval widened by the dispersion between
episodes~\citep{wedderburn1974quasi,mccullagh1989glm}, the device study 5 registers
(\S\ref{sec:s5design}). It was computed in the comparison of intervals only, where it
restores 0.952 in back-office and 0.994 in infra, at 2.5 to 3 times the width, and
gives 0.894 on a pool of longer episodes. It does not repair loss whose rate is drawn
once for an episode and shared by all three of its logs. In the comparison of intervals
that loss was run with the end-state diff shown every operation, 1000 replicates a
cell: the wider interval covers 0.495 in back-office and 0.174 in infra, against 0.432
and 0.124 for the plain one. The ``session quality'' row of Table~\ref{tab:earlier} is
another cell of the same loss, in which the end-state diff sees only the back-office
operations it can see, on other loss seeds and 500 replicates: the plain interval
covers 0.194 there, and the wider one was not computed on it.

\paragraph{Where a counter is the better instrument.} The lower block of
Table~\ref{tab:earlier} sets a tail-corrected sequence counter beside the three-log
estimate. Where the numbers survive only on the records that survive, the gaps give the
losses inside the sequence and the lost tail has to be estimated. When loss fell on
records one by one, both covered above the bar and the counter's interval was the
narrower. When loss fell harder on the operations one log cannot see, the counter kept
its coverage and the estimate lost it. The counter fell under the bar when all three
logs of an episode shared one loss rate, when records were lost in blocks, and when a
twentieth of the actions bypassed the gateway and took no number. So the counter is the
better instrument where it sees every action and records are lost independently of one
another, and it fails where loss is shared, comes in blocks, or an action can go round
it.

All loss in these four studies was simulated, and in the first two one gateway saw
every action. Study 5 removes both.

\section{Study 5: design}
\label{sec:s5design}

Study 5 asks three questions. Q1: does the interval
cover the true count when records are lost by real telemetry components, in blocks, to
both planes at once (criterion P1)? Q2: is the estimate closer to the true count than
counting sequence gaps, and than counting the union of the logs (P2, against each)? Q3
asks the same without the end state, and was predicted to fail.

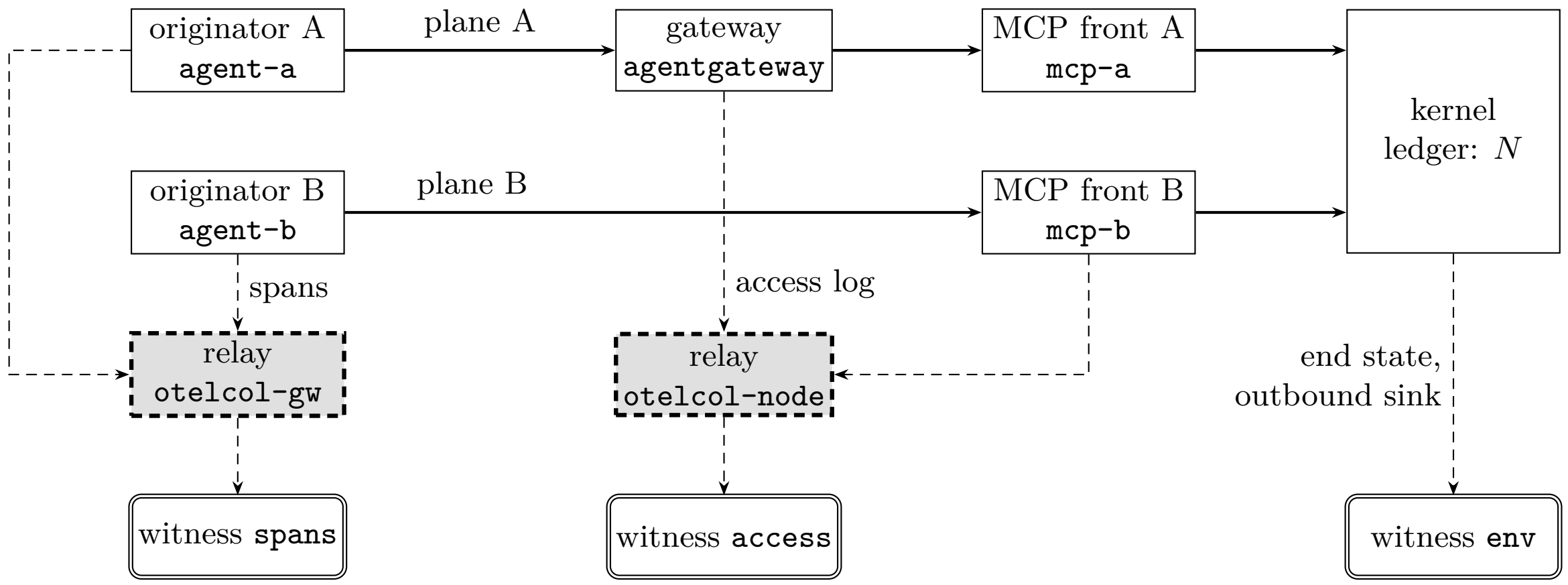
\begin{figure}[!tb]
\centering\footnotesize
\begin{tikzpicture}[x=12mm, y=8mm, >={Stealth[length=4pt]},
  box/.style={draw, align=center, minimum width=21mm, minimum height=8mm, inner sep=2pt},
  killed/.style={box, densely dashed, very thick, fill=black!12},
  wit/.style={box, rounded corners=3pt, double},
  call/.style={->, thick},
  tele/.style={->, densely dashed}]
\node[box] (oa) at (0,4) {originator A\\\texttt{agent-a}};
\node[box] (ob) at (0,2) {originator B\\\texttt{agent-b}};
\node[box] (gw) at (4,4) {gateway\\\texttt{agentgateway}};
\node[box] (fa) at (7,4) {MCP front A\\\texttt{mcp-a}};
\node[box] (fb) at (7,2) {MCP front B\\\texttt{mcp-b}};
\node[box, minimum height=24mm] (k) at (10,3) {kernel\\ledger: $N$};
\node[killed] (rs) at (0,0) {relay\\\texttt{otelcol-gw}};
\node[killed] (rl) at (4,0) {relay\\\texttt{otelcol-node}};
\node[wit] (ws) at (0,-2) {witness \texttt{spans}};
\node[wit] (wa) at (4,-2) {witness \texttt{access}};
\node[wit] (we) at (10,-2) {witness \texttt{env}};
\draw[call] (oa) -- node[above] {plane A} (gw);
\draw[call] (gw) -- (fa);
\draw[call] (ob) -- node[above, pos=0.2] {plane B} (fb);
\draw[call] (fa.east) -- (k.west |- fa.east);
\draw[call] (fb.east) -- (k.west |- fb.east);
\draw[tele] (ob) -- node[right] {spans} (rs);
\draw[tele] (oa.west) -- ++(-1,0) |- (rs.west);
\draw[tele] (gw) -- node[right, pos=0.8] {access log} (rl);
\draw[tele] (fb.south) |- (rl.east);
\draw[tele] (rs) -- (ws);
\draw[tele] (rl) -- (wa);
\draw[tele] (k) -- node[left, align=right] {end state,\\outbound sink} (we);
\end{tikzpicture}
\caption{The study-5 rig. Solid arrows are tool calls: plane A through the gateway,
plane B straight to its MCP front, both to one kernel whose ledger gives the true count
$N$ and which no witness reads. Dashed arrows are what the witnesses are made from. Each
originator sends its spans to one relay; the access logs of the gateway (plane A) and of
MCP front B (plane B) are written to files by the run's driver, not drawn, and those
files are tailed by the other; each relay forwards to a collector, not
drawn, that writes the witness file. \texttt{env} is read from the kernel's end state and
outbound sink when the run is over. The two relays, shaded with a dashed border, are the
components \texttt{s1} kills together and restarts; no originator is killed in
\texttt{s1}. Every witness is split by plane before the tables are summed. Layout from
\texttt{rig/compose.yaml}.}
\label{fig:rig}
\end{figure}

\paragraph{The rig.} Two originators replay recorded sessions over MCP and share no
counter. Plane A goes through an unmodified open-source gateway, \texttt{agentgateway},
and plane B straight to an MCP front. One kernel service keeps the ledger. Each plane
has three witnesses. \texttt{spans} are the originator's own OpenTelemetry spans, sent
through a relaying collector; a batch is given up when the relay is away past the SDK's
retries. \texttt{access} is the receiver's access log, tailed by a second relaying
collector; lines written while that relay is away are skipped. \texttt{env} is the end
state and the outbound sink, each row with its last writer; an action overwritten or
undone later leaves nothing in it. Every SDK and collector setting is its default, no
code in the repository drops a record, and only the two relays and the originators are
ever killed. Both planes' spans pass one relay and both planes' access logs the other.
\texttt{spans} and \texttt{access} are linked by trace id and JSON-RPC id, both stamped
by the client library where no model output reaches them; \texttt{env} attaches by what
the action was. Figure~\ref{fig:rig} draws the layout.

\paragraph{Sessions and runs.} Sessions were recorded once from two models
(Table~\ref{tab:pools}), ticket by ticket, on a task mix in which later tickets correct,
revert, delete or rewrite what earlier ones did, which is what makes the end state blind
to part of the work. The pools were recorded before the registration was frozen, because
their hashes are part of it, and no run replayed them before it. A run replays 24
sessions drawn without replacement from one pool, with session starts spread
over 100 s, each call made when the recording made it. The plan is 420 runs in three
conditions (Table~\ref{tab:conditions}). \texttt{s1}, the primary condition, kills both
relays together at a time drawn from 8 to 90 s and restarts them 15 to 45 s later. The
analysis ran once, after every run had a folder and a hash of each folder was committed.

\begin{table}[!tb]
\centering\small
\input{tables/study5_pools}
\caption{Study 5: the pools of recorded sessions, and coverage by pool with the floors,
on the runs left (\S\ref{sec:s5results}).}
\label{tab:pools}
\end{table}

\begin{table}[!tb]
\centering\small
\input{tables/study5_conditions}
\caption{Study 5: the three registered conditions, the runs set aside by reason, and
what each condition lost.}
\label{tab:conditions}
\end{table}

\paragraph{The registered estimator.} \texttt{ll\_sa\_hom} sums the two planes' tables
and names one model in advance. Its point is the plain two-list estimate between any
telemetry and the end state; what it adds is the interval. The simple alternative is
Chapman's estimate with a Wald interval on the same two lists, from the variance Seber
gives for it~\citep{seber1982estimation}, \texttt{chapman\_te},
which the registration reports beside it with no criterion; \S\ref{sec:s5results} sets
the two side by side, as a description. Write $y_h$ for the number of a run's actions with
capture history $h$, the set of witnesses that hold the action, with S for
\texttt{spans}, A for \texttt{access} and E for \texttt{env}, summed over the two planes.
The cell $y_\varnothing$ is not observed, and $n=\sum_{h\neq\varnothing}y_h$ is the count
seen. The point is Lincoln--Petersen between two lists, any telemetry (spans or access)
and the end state:
\begin{equation}
\hat N = n+\hat z_0,\qquad
\hat z_0=\frac{y_{\mathrm{E}}\,\bigl(y_{\mathrm{S}}+y_{\mathrm{A}}+y_{\mathrm{SA}}\bigr)}
{y_{\mathrm{SE}}+y_{\mathrm{AE}}+y_{\mathrm{SAE}}},
\label{eq:point}
\end{equation}
the count seen, plus the actions seen by the end state alone times those seen by
telemetry alone, over those seen by both. That $\hat z_0$ is the maximum-likelihood
missing cell of the Poisson log-linear model whose only interaction is between spans and
access. With $s_h$, $a_h$ and $e_h$ equal to 1 when $h$ holds S, A and E and 0 when not,
\begin{equation}
\log\mu_h=\beta_0+\beta_{\mathrm{S}}s_h+\beta_{\mathrm{A}}a_h+\beta_{\mathrm{E}}e_h
+\beta_{\mathrm{SA}}\,s_h a_h,\qquad
\hat z_0=\hat\mu_\varnothing=e^{\hat\beta_0},
\label{eq:model}
\end{equation}
in which the two telemetry logs may lose the same records and the end state is taken to
miss actions independently of both. The interval is that model's profile likelihood over
whole counts $N\ge n$~\citep{cormack1992interval}. With the table completed by $y_\varnothing=N-n$ and
$\hat\mu_h(N)$ the model's fit to it,
\begin{equation}
\ell(N)=\log N!-\log(N-n)!+\sum_{h}y_h\log\frac{\hat\mu_h(N)}{N},
\qquad
\mathrm{CI}=\Bigl\{N\ge n:\ \ell(N)\ge\ell(\tilde N)-\tfrac12\,\hat\phi\,\chi^2_{1}\Bigr\},
\label{eq:profile}
\end{equation}
where $\tilde N$ is the whole count that maximises $\ell$, searched from $\hat N$,
$\chi^2_{1}$ is the 95\% point of the chi-squared distribution on one degree of freedom,
and the two ends are found by bisection either side of $\tilde N$. The point is a real
number and the ends are whole counts. The cutoff is scaled by $\hat\phi$, the dispersion
between the run's tables by session and plane~\citep{wedderburn1974quasi,mccullagh1989glm}.
Take the $m$ such tables that hold an action, $y_{ij}$ the count of table $i$ in observed
cell $j$, $n_i=\sum_j y_{ij}$, $p_j=\sum_i y_{ij}/\sum_i n_i$, and $J$ the number of cells
with $p_j>0$:
\begin{equation}
\hat\phi=\max\Bigl\{1,\ \frac{1}{(m-1)(J-1)}\sum_{i=1}^{m}\sum_{j:\,p_j>0}
\frac{(y_{ij}-n_i p_j)^2}{n_i p_j}\Bigr\},
\label{eq:phi}
\end{equation}
the Pearson statistic of homogeneity over its degrees of freedom, never under 1, and 1
when $m<2$ or $J<2$. It uses no log-linear model, so a dependence between logs is not
read as dispersion. Nothing is selected. The search stops at a cap of $20n$, and the
estimator returns no upper end in four cases. If no action is in both lists, the
denominator of (\ref{eq:point}) is 0 and nothing ties the end state to the telemetry; if
$\hat N>20n$, the point is past the cap; in both it returns no point, a lower end of $n$
and no upper end. If $\ell$ is still at or above the cutoff past $20n$,
the point stands and the upper end is infinite. And if the iterated fit of
(\ref{eq:model}) fails at a count the search visits, it returns no point and no upper end.

\paragraph{Baselines and criteria.} \texttt{seq\_gap\_tail} reads the JSON-RPC ids on
the delivered telemetry of each session and plane, counts the ids missing up to the last
one seen, converts requests to actions at the share of seen calls that were actions, and
adds the expected lost tail. It is told each session's first id and is not shown the end
state. P1 passes if the Wilson 95\% lower end~\citep{wilson1927probable} of coverage over
the \texttt{s1} runs left
is at or above 0.90 and under 0.05 of intervals are unbounded; an interval with no upper
end counts as covering unless its lower end is above the truth. P2 passes against a
baseline if the exact one-sided 97.5\% lower bound on the share of runs in which the
point is strictly closer to the true count is above 0.5, a tie counting as a loss. A
fail reads ``missed'' when the interval excludes meeting the criterion and ``not shown''
otherwise. Two gates come first: fewer than 270 \texttt{s1} runs left is
``underpowered'', and a mean share of actions in no log under 0.02, or telemetry that
loses under 0.01 of a plane's actions, is ``not exercised''.

\paragraph{Integrity rules.} Before any test, a run is set aside if any of these holds:
a stress event did not happen; an originator exited wrongly, left no report or could not
replay a session; a witness file is missing or empty; a service that should have been up
was not; the copy of an access log is incomplete; a plane's receiver did not get every
planned call; a telemetry record has no session or matches no ledger entry; a relay
delivered no line of a plane's access log; a receiver and the ledger disagree on the
actions; a call was made more than 0.25 s late; an originator was not ready at the
start; the containers' clock is more than 0.5 s from the driver's; the driver did not
finish; or the run is not a registered run. Two of these can follow from the loss under
study itself, an empty witness file and a relay that delivered no access line: a run
lost that way is a run with extreme loss left out. So the registration also has P1's
coverage reported with every planned \texttt{s1} run that is not in the test counted as
not covered, beside the criterion and with no criterion of its own. A run is made once
and a run set aside is not replaced.

\section{Study 5: results}
\label{sec:s5results}

\paragraph{The runs.} All 420 runs were made. The integrity check set aside 34 of them
before any test: 2 of 60 in \texttt{s0}, 23 of 300 in \texttt{s1} and 9 of 60 in
\texttt{s2} (Table~\ref{tab:conditions}). Neither gate fires: 277 \texttt{s1} runs are
left against a bar of 270, and the mean share of a run's actions in no log is 0.068. An
\texttt{s1} run has 316 to 353 actions.

\paragraph{P1.} The criterion as registered is met on the runs left:
\texttt{ll\_sa\_hom} covered 269 of 277 \texttt{s1} runs (Wilson 0.944 to 0.985) and no
interval was unbounded, so both clauses read ``pass''. 23 of the 300 planned \texttt{s1}
runs were set aside on integrity, and with those counted as not covered it is 269 of
300, Wilson 0.857 to 0.926, which straddles the 0.90 bar and reads ``not shown''. The
registration gives that second figure no criterion and puts it beside the verdict; it
had predicted that the coverage clause would not be met
(Appendix~\ref{app:registration}). Three
caveats are fixed by the registration. The bound treats runs as independent, and the 100
planned runs of a pool draw 24 of the same 72 sessions (below). This is one rig. And an
interval with no upper end
would have counted as covering; none occurred in \texttt{s1}, so coverage over bounded
intervals alone is the same count. Of the eight runs not covered, seven intervals end
below the true count, by 1 to 5 actions, and one starts 9 above it. The pilot's misses
are reported beside the verdict, as registered: the interval missed in two of its 20
runs on recorded sessions and one of its 24 on scripted ones
(Appendix~\ref{app:registration} names them).

\begin{figure}[!tb]
\centering
\includegraphics[width=\linewidth]{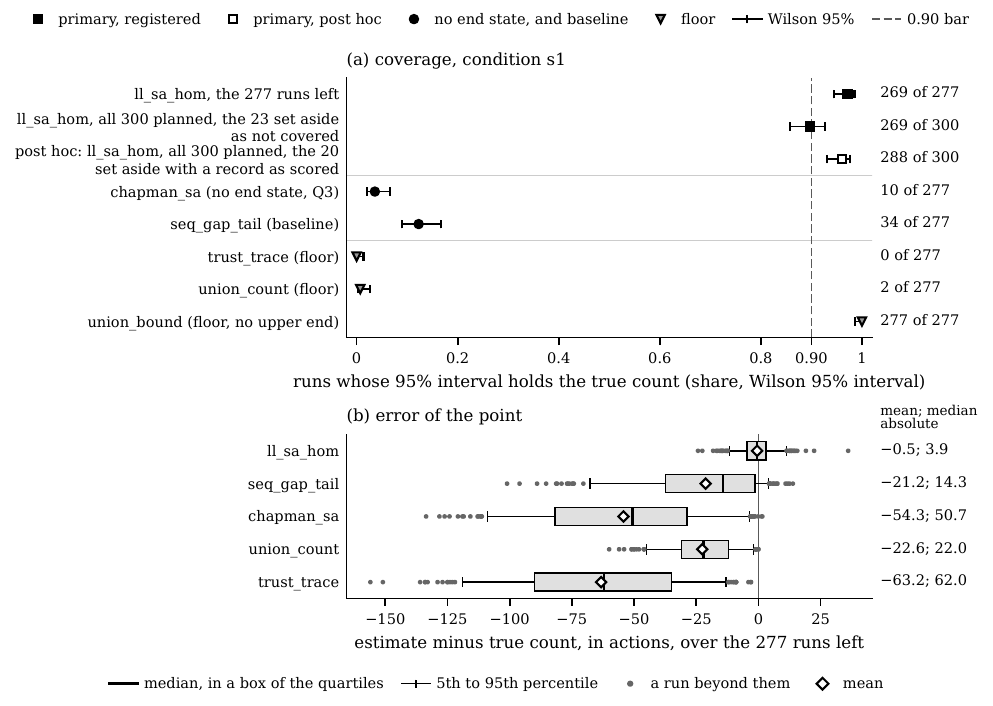}
\caption{Study 5, condition \texttt{s1}. (a) The share of runs whose 95\% interval holds
the true count, with the artifacts' Wilson 95\% intervals, against the 0.90 bar (dashed).
The first three rows are the registered estimator: on the 277 runs left of the 300
planned, 23 set aside on integrity; on all 300 with those 23 counted as not covered, which
reads ``not shown''; and, as an open square, post hoc, with the 20 set aside that left a
full record scored as they were. Below them are the estimate without the end state, the
sequence-gap baseline and the three floors; \texttt{union\_bound} covers every run
because none of its intervals has an upper end. (b) The error of the point in each of the
277 runs: a box is the quartiles and the median, the whiskers are the 5th and 95th
percentiles, a dot is a run beyond them and the diamond is the mean. At the right are the
mean error and the median absolute error, in actions.}
\label{fig:s5results}
\end{figure}

\paragraph{The runs set aside.} None of the 23 is of the two kinds the registration says
can follow from the loss under study, an empty witness file and an access log with no
line delivered. Two things were read about them post hoc, with no criterion
(Appendix~\ref{app:aside}). The first is why they were set aside: the power log of
the laptop that made the runs puts most of them beside a sleep of the machine or in a
stretch on battery. Of the 34 runs set aside over the
three conditions, 8 had a call 20 s or more late or a clock 20 s or more off: 3 were
made before the retained log begins, and each of the other 5 sits beside a sleep of the
machine, as each of the 6 sleeps the log records while runs were being made sits beside
one of them. Another 21 had a call late, or an originator slow to start, by less than that (the
remaining 5 left no record or delivered no line of an access log). Of the 21, 17 are among the 58 runs started
between the log's last reading on battery, at 11\% charge, and the moment external power
was plugged in, against 3 among the 314 runs before and 1 among the 45 after. In that
stretch runs with no stress were set aside as well (2 of 9 in \texttt{s0}, against 11 of
41 in \texttt{s1} and 4 of 8 in \texttt{s2}; Table~\ref{tab:host}), so the lateness there
did not need the loss under study. What slowed the machine was not measured: the log
gives sleeps and the power source, not the processor's speed. The second is what the estimator returned on them. 20 of the 23 left a
full record and 3 left none. With those 20 scored as they were, the interval held the
true count in 19, which with the registered count is 288 of 297, and 288 of 300 with the
3 that left no record counted as not covered (Wilson 0.931 to 0.977). A run that broke
the timing rule is not a run of the design, so this is a description and not a test, and
the two registered figures stand as they are.

\paragraph{Dependence between runs.} The usual repair when units share material is to
resample the shared thing, here the sessions, and this design does not allow it. Each
pool holds 72 sessions and each run draws 24 of them, so a session is on average in a
third of its pool's runs, no run is independent of another run of its pool, and a run's
outage was drawn once and stays with the run. Post hoc, the Wilson lower end keeps the
bar up to a design effect of 4.05. The dependence can be estimated under a model in
which a run's outcome is its pool's mean plus a term for each session it replays: the
covariance of two runs then grows with the sessions they share, 8.0 on average over the
12652 pairs of runs of one pool. For whether a run's interval missed, which is what the
bound counts, the share of variance tied to sessions is estimated at 0.063 (permutation
$p=0.244$), a design effect of 2.92, at which the lower end is 0.915. That estimate is
imprecise: its approximate upper limit is a design effect of 7.51, past what the bound
bears. So these runs neither show dependence nor exclude enough of it to matter, and
settling it takes sessions that no two runs share (\S\ref{sec:conclusion}).

\paragraph{Two descriptions made after the verdicts were read.} Both are post hoc and
have no criterion, and no verdict depends on either. They are over the 277 \texttt{s1}
runs in the test, 23 of the 300 planned set aside. The first is the pooled figure on
the estimate's assumption, that what both telemetry logs lose is as visible in the end
state as the rest. Both telemetry logs lost 15475 actions, of which the end state shows
9223, a share of 0.596; of the other 76617 actions it shows 46294, a share of 0.604.
The counts are from a table the registration labels exploratory, which is published and
used for no claim for or against the estimate by band, pool or outage length. They are
pooled over runs that share sessions: the two shares are an average, say nothing about
any one run, and are not a test of the assumption. The second is the dependence each run
would have needed. The odds ratio between the two lists of (\ref{eq:point}), any
telemetry and the end state, at which the two-list count equals the true count is
$(N-n)/\hat z_0$, the actions in no log over the number (\ref{eq:point}) adds;
independence is 1. It is defined in 274 of the 277 runs; in the other 3 the end state
held no action the telemetry lacked. Its median is 1.039 and its quartiles are 0.828
and 1.288; 137 runs are within a factor of 1.25 of 1 and 201 within a factor of 1.5.
That describes how far each run sat from independence, in the spirit of the work on
dependence between lists cited in \S\ref{sec:related}. It is none of those papers'
methods, and nothing is computed from them.

\paragraph{Three further descriptions, post hoc.} These were computed for this version,
after the verdicts were read and knowing the results. They have no criterion, no verdict
depends on any of them, and all describe the 277 \texttt{s1} runs of the test, 23 of the
300 planned set aside. \emph{Where the error comes from.} In every run the error of the
point of (\ref{eq:point}) is exactly the number of actions both telemetry logs lost,
times the share of them the end state shows less the share it shows of the rest, over
that second share. The end state does show actions unevenly by place in a session: 0.441
of those in the first fifth of their session and 0.807 of those in the last. A run's
sessions start spread over 100 s, so one outage falls on a different part of each, and
the lost actions sat across the fifths much as the rest did (0.240 of them in the first
fifth against 0.222 of the rest, and 0.158 against 0.170 in the last). Of the pooled
difference between the two shares, $-0.008$, place accounts for $-0.006$ and a
difference within a fifth for $-0.002$. That holds on average and not run by run: place
carries 0.227 of the variance of a run's difference and the part within a fifth 0.773.
In each of the eight runs not covered the difference was more than 1.96 times the
deviation it would have by chance, as it was in 14 of the 268 covered runs that lost an
action, and their intervals were narrow (median width 16 actions against 32). Place has
the sign of the difference in all eight and is the smaller part in 7. With an interval
shifted by the place part and not fitted again, 4 of the 8 would hold the true count and
4 covered runs would not; shifted by the part within a fifth, all 8 would. So the spread
of session starts evens out where the lost actions sit, and what is left of each miss is
mostly not place. A workload whose sessions start together, or one session, has no such
averaging. By arithmetic on the pooled shares, which is no run: were the lost actions all the
actions of the first fifth of their sessions, shown at that fifth's rate, and the rest
the actions of the other four fifths, the point would be 0.32 of an action low for each
lost action, and with the last fifth 0.44 high, the failure of \S\ref{sec:earlier}.
Those two figures rest on that choice of what the rest is: with the rest held at its
pooled rate of 0.604, and not emptied of the fifth that was lost, the same arithmetic
gives 0.27 low and 0.34 high. \emph{An interval under
bounded dependence.} The repository's \texttt{anchor\_bounds} keeps the point of
(\ref{eq:point}) and takes the odds ratio between the end state and the telemetry to
lie between 1 and a stated $\Gamma$, an assumption of the kind the bounds
of~\citep{sun2020bounds} make and not their method. At $\Gamma=1$ its interval held the
true count in 250 of the 277 runs (0.903) at a mean width of 23.9 actions, against 32.6
for the registered interval. At 1.25, with a width of 32.1, it held 263 (0.949); at 1.5,
with a width of 40.3, it held 271 (0.978), 6 of the 8 runs the registered interval
missed among them; and at 2 the same 271 at a width of 56.6. Raising $\Gamma$ from 1.5 to 2 only raises the upper
end, and that does not help the 6 runs still missed: 4 have an interval above the true
count, where an odds ratio of 1 or more does not reach, and in the other 2 the end state
held no action the telemetry lacked, so the count that (\ref{eq:point}) adds is zero,
$\Gamma$ multiplies zero, and at 2 the interval is still below the true count (no
larger $\Gamma$ was run). Of the 6, 4 are
runs the registered interval covered; this interval is not widened by the dispersion
between sessions. The four values of $\Gamma$ were chosen after the results. At the
registered interval's width the bound covered fewer runs. \emph{The limit on the design
effect.} The upper limit of 7.51 on the design effect (above) takes the permutation
distribution as normal. Read from the same 10000 permutations with no such
approximation it is 7.59 by the null's 95th percentile and 7.41 by inverting the test.
A bootstrap of runs within a pool, itself approximate, gives 8.31, at which the Wilson
lower end is 0.850, and 0.338 of its resamples exceed the 4.05 the bound bears. All four
limits are past 4.05, so the reading above stands as it is.

\begin{table}[!tb]
\centering\small
\input{tables/study5_s1}
\caption{Study 5, condition \texttt{s1}: every registered method, the baselines and the
floors, with their coverage in \texttt{s0} and \texttt{s2}. With the 23 runs set aside
counted as not covered, P1's coverage is 269 of 300 (``not shown'').}
\label{tab:s1}
\end{table}

\paragraph{P2.} It passes against both baselines, on the same 277 of 300 planned runs.
The point was closer to the true count than \texttt{seq\_gap\_tail} in 223 runs (exact
lower bound 0.753) and than \texttt{union\_count} in 261, with 3 ties (0.908). The two
comparisons do not weigh the same. The union is never above the true count and the
estimate is the union plus a non-negative amount, so against \texttt{union\_count} the
comparison shows that the amount added is the right order, not that it is accurate. The
union was already exact in 2 runs, where the estimate can only tie or lose. The
informative comparison is the one with \texttt{seq\_gap\_tail}, and it compares
information as much as method: \texttt{seq\_gap\_tail} reads request ids on delivered
telemetry and is never shown the end state, the witness that Q3 shows carries the
signal. As counts (Table~\ref{tab:s1}), \texttt{seq\_gap\_tail} covered 34 of the 277
runs, \texttt{union\_count} 2, \texttt{trust\_trace} none, and \texttt{union\_bound}
every run while bounding none.

\paragraph{Q3, without the end state.} \texttt{chapman\_sa} is Chapman's estimate
between \texttt{spans} and \texttt{access} alone, on the same 277 of 300 planned runs.
Its coverage clause reads ``missed'': 10 of 277 runs covered (Wilson 0.020 to 0.065).
Its unbounded clause reads ``pass'': no interval was unbounded. P2 reads ``missed''
against both baselines: 4 wins and a tie against \texttt{seq\_gap\_tail}, and 3 wins
against \texttt{union\_count}. Its mean error is $-54.3$ actions. The two telemetry logs
lose to the same outage, so their overlap says little about what both missed.

\paragraph{Control and the heavy condition.} 58 of the 60 \texttt{s0} runs are in the
test (Table~\ref{tab:s1}, last columns). No interval lies above the true count, and every
estimator's point is the count in every run that has a point. One run has none. In
\texttt{s5-s0-021}, where no log lost anything, \texttt{ll\_sa\_hom} did not converge
and returned no point and no upper end, with the client library's ids and with no id
trusted. By the registered rule that interval counts as covering. An estimator that
returns nothing on a complete log is defective. On a table with five empty cells the iterated fit loses rank where $N$ is the
count seen. Post hoc, the library's version of the estimator falls back to the model's
closed form there; the registered artifacts are not regenerated and no verdict changes.
\texttt{s2}, the heavy condition, has no criterion and no prediction. 51 of its 60 runs
are in the test, and \texttt{ll\_sa\_hom} covered 49 of them (Wilson 0.868 to 0.989)
with none unbounded. The union was exact in 18 of those
runs. The point was closer than \texttt{union\_count} in 25 of the 51, which is under
half, with 15 ties, and closer than \texttt{seq\_gap\_tail} in 28, with 15 ties.

\paragraph{Beside the primary estimator.} The registration reports seven other estimates
with no criterion, and none stands in for the primary one (Table~\ref{tab:s1}). With no
id trusted the primary estimate gave the same counts. \texttt{chapman\_te} is Chapman's
estimate on the same two lists, any telemetry and the end state, with a Wald interval.
Its point did as well as the registered one: closer than \texttt{seq\_gap\_tail} in
224 runs against 223, with a median absolute error of 4.0 actions against 3.9. Its
interval covered 250 of the 277 runs and is the narrower (Table~\ref{tab:s1}, width).
So on these runs what the registered estimator adds to a plain Chapman interval on the
same two lists is coverage, at a cost in width. That comparison is descriptive: no
criterion was registered for \texttt{chapman\_te}. Also post hoc: against the larger of the
gap count and the union, which does read the end state, the point was closer in 219 of
the 277 runs (lower bound 0.738). Coverage by pool is in
Table~\ref{tab:pools}, and the registered exploratory tables, by outage length and by
the number of actions both telemetry logs lost, are in the study's two artifacts and
in no table here. This paper makes no claim, for or against the estimate, from a
table by band, pool or outage length: it concludes about \texttt{s0}, \texttt{s1} and
\texttt{s2} only.

\section{The six conditions}
\label{sec:when}

Six conditions separate the case in which the count was recovered from the cases in
which it was not. They held together in study 5's primary condition, which tested none
of them in isolation; each is set here against the study in which it did not hold. None
can be read off an estimate, because the estimate does not detect its own failure
(\S\ref{sec:earlier}): they are properties of the deployment, established by reviewing
its design.

\begin{enumerate}\setlength\itemsep{1pt}\setlength\parskip{0pt}
\item \textbf{No counter can be read.} In study 5 the two originators shared none, and
the sequence baseline, with only the ids on delivered records, covered 34 of 277 runs.
The rig's single kernel is itself a census at the sink. It stands for downstream systems
whose logs are not collected, which the study stipulates and does not show.
\item \textbf{Records are linked exactly.} Matching by effect passed study 1 at its
registered seed, gave 0.893 over 30 seeds, and failed study 2. An id stamped by
the harness or the client library, outside the model's reach, did not fail. In
study 5, where few actions are alike, linking with no id trusted gave the same counts.
\item \textbf{Every channel is itemised by at least two logs, and a log does not lose a
batch as one.} With a batch tool that no log itemised, coverage fell under the bar on
both models (Table~\ref{tab:earlier}). With the scripts parsed and their lines lost together it was 0.335 to 0.610. Study 5 has
no batch tool: it assumes this condition.
\item \textbf{One witness is of another kind.} The end state misses what was overwritten
or undone, and the telemetry what an outage took. With both, the estimate is the case
above. With the two telemetry logs alone it covered 10 of the same 277 runs. In
\texttt{s1} both relays are killed together by design, so that result does not speak to
telemetry logs whose collectors fail independently.
\item \textbf{What the logs lose together is as visible to that witness as the rest.}
This is the estimate's assumption. Simulated loss that broke it took coverage to 0.187
(0.254 on the loss seeds of Table~\ref{tab:earlier}) with no sign in the output. Study 5
does not test it within a run, and its pilot saw the end state show actions unevenly by
kind and by place in a session. Its sessions start spread over 100 s, which evens out
where in a session the lost actions sit; sessions that start together, or one session,
would not have that (\S\ref{sec:s5results}, post hoc). Post hoc and pooled over the 277 runs, the end state
shows 0.596 of what both telemetry logs lost and 0.604 of the rest, an average that is
not a test, and the odds ratio a run would have needed between its two lists has
quartiles 0.828 and 1.288 (\S\ref{sec:s5results}).
\item \textbf{The interval allows for loss in blocks.} In the rig a relay's outage takes
one window from every session on both planes, and the registered interval is widened by
the dispersion between sessions. Under simulated blocks the plain interval covered 0.559.
\end{enumerate}

\paragraph{Applying the conditions.} A review asks which processes originate actions,
which routes an action can take, which component writes each log and by which pipeline
it arrives, and what the logs can lose together. The repository's tool has a
\texttt{lint} command that reads a map of the logs, and no log, and checks part of
this: that each plane has at least two witnesses, that they do not all arrive by one
pipeline, and that no two blocks of a witness with no id read the same rows. It checks
what the map says. It cannot check that the map is true, that its planes are every route an action
can take, or that the witnesses lose records independently.

\section{Limitations}
\label{sec:limits}

\begin{itemize}\setlength\itemsep{0pt}\setlength\parskip{0pt}
\item \textbf{One rig.} Study 5 is one rig, one layout and one outage shape per
condition, on a run of 113 to 133 s with one outage in it, so it is not a general map
and says nothing about a deployment's rate of loss. The registered result is on 277 of
the 300 planned runs, with 23 set aside (\S\ref{sec:s5results}). The bar is 0.90, five
points under the interval's level, so the study does not show coverage at 95\%. Nothing
here says at which outage lengths, on which pools or at which amounts of loss the
estimate works or fails: the registration bars that reading and the study is not sized
for it.
\item \textbf{Runs share sessions.} The bound on coverage takes runs as independent.
The dependence was estimated post hoc and too loosely to settle
(\S\ref{sec:s5results}).
\item \textbf{No adversary in study 5.} Nothing in it deletes or alters a record on
purpose. The link id is stamped by the client library in
the originator's process, and a hostile process can stamp what it likes. Studies 1 to 4
ran benign agents and one weak adversary prompt.
\item \textbf{Replay, not live agents.} The sessions come from two model families served
by one provider and are replayed on a clock, so a model that would have reacted to a
lost reply is not modelled. Every statement about study 5 is conditional on the 216
sessions recorded.
\item \textbf{The planes are summed.} If their loss or their hiding differ, the pooled
estimate is biased and nothing checks for it. The dispersion that widens the interval
comes from tables that share one outage and cannot see what every table shares.
\item \textbf{Simulated loss in studies 1 to 4.} Their logs were complete and loss was
drawn on them afterwards, and their repairs are post hoc.
\item \textbf{Design choices made by the author.} The task mix decides what the end
state hides, and its weights were set to hide about two fifths of the actions. The
estimator and the primary condition were chosen after a pilot on the same rig. The
environments and the bars are the author's, and no study applies a multiplicity
correction.
\end{itemize}

\section{Conclusion}
\label{sec:conclusion}

The actions that no log of an agent recorded can be counted from several incomplete
logs, and five pre-registered studies say when. In the one live case tested, a rig
whose telemetry loses records by itself, \texttt{ll\_sa\_hom} covered 269 of the 277
\texttt{s1} runs left (the criterion as registered is met). 23 of the 300 planned runs
were set aside on integrity; with those counted as not covered it is 269 of 300, whose
Wilson interval straddles the bar and reads ``not shown'', and post hoc, with the 20
that left a full record scored as they were, it is 288 of 300. The point was closer to
the true count than counting gaps or counting the union, and without the end state the
interval held the true count in 10 of the same 277 runs. The same estimate fails, with
nothing in its output to show it, when records are not linked exactly, when an action
can take a route no log itemises, when a log loses a batch as one, and, post hoc, when
loss depends on the action or comes in blocks the interval does not allow for.

What separates the two is six conditions of the deployment, which a design review can
check and the estimate cannot. Four have a failure in the earlier studies, the witness
of another kind has its failure inside study 5, and the first, that no counter can be
read, is a premise. The next study needs sessions that no two runs share, relays that
fail independently of each other, a condition in which what the end state hides depends
on what the logs lose, and a second rig.

\section*{Disclosures and availability}

The code, the analysis scripts and the text were written with an AI coding assistant,
under the author's direction and review. The author works on a commercial AI
gateway; nothing from that product is used, every component is open source and all data
is synthetic. The code, the five registrations, the decision log and every artifact a
number here comes from are at \url{https://github.com/caiotheodoro/lacuna}. The capture
histories and study 5's runs, without the transcripts, are at
\url{https://huggingface.co/datasets/caiotheodoro/lacuna-capture-histories}. A first
account of studies 1 to 4 was published on 31 October 2026 and of study 5 on 16 November
2026.

\bibliographystyle{plainnat}
{\footnotesize\setlength{\bibsep}{0pt}\bibliography{refs}}

\appendix

\section{Registrations, predictions and deviations}
\label{app:registration}

\paragraph{The freeze.} Each registration is a commit that predates the runs it governs.
Study 5's is a signed tag, \texttt{prereg-study5}, pushed with a release before the
first confirmatory run. No copy of it was deposited outside GitHub, though an
earlier decision had asked for one before the first paid call, so GitHub's is the only
clock on it that the author does not control. Its text was prepared and tagged by the AI
assistant on the author's instruction, and the author read its final text after the
runs.

\paragraph{Predictions.} Studies 2 to 5 registered predictions of their own results.
Study 2 predicted that its cell would fail, and it failed. Study 3 predicted that the
agent would run 0.05 to 0.30 of its actions through the batch tool, and it ran 0.856.
8 of study 4's 12 predictions held in full. Study 5 predicted that P1 would not be met
on its coverage clause. The basis was the pilot: 18 of 20 runs covered on recorded
sessions (Wilson 0.699 to 0.972) and 23 of 24 on scripted ones (0.798 to 0.993), which
with a flat prior gave a pass a probability of 0.165 and 0.494.
Table~\ref{tab:predictions} sets each of study 5's predictions against its result. Six
were right. The prediction that P1 would not be met on its coverage clause was wrong by
the registration's rule: the clause passed on 277 of the
300 planned runs, 23 set aside, and it is not shown with those counted as not covered
(\S\ref{sec:s5results}). Its gloss was ``with all 300 left, fewer than 281
covered'', and not all 300 were left.

\begin{table}[!htb]
\centering\small
\input{tables/study5_predictions}
\caption{Study 5: the registered predictions and which were right.}
\label{tab:predictions}
\end{table}

\paragraph{The pilot's misses.} On recorded sessions: \texttt{ptr-ds-restart-6} (325
actions, interval 326 to 342) and \texttt{ptr-ds-restart-7} (330, interval 326 to 327).
On scripted sessions: \texttt{pt3-node-restart-0} (372, interval 367 to 369).

\paragraph{Deviations, and what was seen before the analysis.} A draft of study 5's
registration split the primary condition in two by outage length; the split was
withdrawn before the freeze, because its line was drawn after the pilot's misses, and
\texttt{s1} is the whole range. The launcher stopped once, when the container engine
quit on the machine, and was started again, as the registration provides: the folder
that stopped it stays as it is and is set aside, and no frozen file changed. The frozen
text has the registration's record ``committed with the times''; the record was
committed alone and the times entered in the log beside it, a reading of an ambiguity
made before any run. Before the seal the launcher's console had shown each run's exit
status, so the number of runs that failed integrity was known before the analysis, with
the file names of four run folders; no run's content and no estimate was seen. The raw
runs are deposited with the repository's release \texttt{v1.0.0rc1}, as the registration
has it, with a hash of each run folder in the repository.

\section{The runs set aside}
\label{app:aside}

Table~\ref{tab:aside} counts the 34 runs set aside by condition, by reason and by place
in the order the runs were made. Table~\ref{tab:host} sets them beside the power log of
the laptop that made them, which is the machine's own record of when it slept, woke and
changed its power source. The retained log begins after the first 3 of the 8 long runs
were made. It gives the power source at each sleep and wake and not in between, and it
does not give the load.

\begin{table}[!htb]
\centering\small
\input{tables/study5_set_aside}
\caption{Study 5: the runs set aside on integrity, by condition and by place in the order
the runs were made. Counts of runs only.}
\label{tab:aside}
\end{table}

\begin{table}[!htb]
\centering\small
\input{tables/study5_host}
\caption{Study 5: the runs set aside, by when they started against the power log of the
machine that made them. Counts of runs only.}
\label{tab:host}
\end{table}

Scoring the 20 runs set aside in \texttt{s1} that left a full record used the registered
code on the sealed folders. The one of the 20 that was not covered has an interval that
ends 1 action below the true count. The median absolute error of the point over the 20
is 4.1 actions, against 3.9 over the 277 runs in the test. The 3 runs with no record are
the one under way when the container engine quit and two whose driver stopped before
the originators were up.

\end{document}

%% file: tables/v1_earlier.tex
\resizebox{\linewidth}{!}{\begin{tabular}{llrrrrr}
\toprule
cell & linkage & \texttt{ll\_bic} & sequence counter & \texttt{trust\_trace} & \texttt{union\_count} & \texttt{union\_bound} \\
\midrule
study 1, back-office & signature & 0.940 (42.0) & -- & 0.000 & 0.000 & 1.000$^{\infty}$ \\
study 2, infra & signature & 0.300 (55.8) & -- & 0.000 & 0.000 & 1.000$^{\infty}$ \\
study 3, batch tool no log itemises & call id & 0.759 (333.6) & 0.000 (32.8) & -- & 0.000 & -- \\
study 3, reference pool, no batch tool & call id & 0.946 (39.0) & 0.954 (26.1) & -- & 0.000 & -- \\
study 4, second model, batch tool & call id & 0.200 (4780.8) & 0.069 (330.5) & -- & 0.000 & -- \\
\midrule
per-record loss & ledger id & 0.952 (38.9) & 0.944 (26.2) & 0.000 & 0.000 & 1.000$^{\infty}$ \\
loss on unseen operations, 1.2 times & ledger id & 0.254 (45.5) & 0.952 (28.6) & 0.000 & 0.000 & 1.000$^{\infty}$ \\
loss in blocks of 12 actions & ledger id & 0.574 (49.1) & 0.000 (19.3) & 0.000 & 0.000 & 1.000$^{\infty}$ \\
session quality shared by all three logs & ledger id & 0.194 (101.8) & 0.480 (25.7) & 0.000 & 0.000 & 1.000$^{\infty}$ \\
5\% of actions bypass the gateway & ledger id & 0.926 (40.8) & 0.002 (26.2) & 0.000 & 0.000 & 1.000$^{\infty}$ \\
\bottomrule
\end{tabular}}
\par\smallskip{\footnotesize Coverage of the 95\% interval (mean width in actions); the bar is 0.90. Upper block: each study's registered cell, 100 replicates at the registered seed (studies 1, 2) or 3000 over 30 loss seeds (3, 4). Lower block: post hoc, the study-1 episodes, 500 replicates a row. The counter numbers an action by its position in the gateway's complete log; only the records it numbered are lost, and an action that bypasses the gateway takes no number. --: not in that artifact. $^{\infty}$: every interval has no upper end. Sources: \texttt{artifacts/thinning/\{confirm,transfer\}.json}, \texttt{artifacts/robustness/bypass-analysis-\{bypass-llm,confirm\}.json}, \texttt{study4-adversary-s4-qwen-bypass.json}, \texttt{baselines.json}.}

%% file: tables/study5_pools.tex
{\footnotesize\setlength{\tabcolsep}{3.5pt}\begin{tabular}{@{}lrrrrrrrr@{}}
\toprule
pool & sessions & actions & hidden & runs left & \texttt{ll\_sa\_hom} & \texttt{trust\_trace} & \texttt{union\_count} & \texttt{union\_bound} \\
\midrule
\texttt{ds} & 72 & 994 & 0.384 & 92 & 89 & 0 & 0 & 92 \\
\texttt{glm} & 72 & 1001 & 0.409 & 94 & 91 & 0 & 0 & 94 \\
\texttt{mix} & 72 & 999 & 0.395 & 91 & 89 & 0 & 2 & 91 \\
\bottomrule
\end{tabular}}
\par\smallskip{\footnotesize Models: \texttt{ds}, \texttt{deepseek-v4-1-flash-1017} on both planes; \texttt{glm}, \texttt{zai-glm-5-3} on both planes; \texttt{mix}, \texttt{deepseek-v4-1-flash-1017} on plane A and \texttt{zai-glm-5-3} on plane B. Actions: ledger actions as recorded. Hidden: one minus the rows the end state holds over the actions. Runs left: of the 100 \texttt{s1} runs planned on each pool. Last 4 columns: runs left whose interval holds the true count, reported as registered with no criterion; a pool is one model pairing on its own ticket seeds, and no conclusion is drawn by pool. Sources: \texttt{pool.json}, \texttt{registered.json} and \texttt{confirm.json} under \texttt{artifacts/study5}.}

%% file: tables/study5_conditions.tex
\begin{tabular}{lrrrr}
\toprule
 & \texttt{s0} & \texttt{s1} & \texttt{s2} & all \\
\midrule
runs planned & 60 & 300 & 60 & 420 \\
runs set aside on integrity & 2 & 23 & 9 & 34 \\
\quad a call made more than 0.25 s late & 2 & 17 & 7 & 26 \\
\quad an originator not ready at the start & 1 & 4 & 1 & 6 \\
\quad the containers' clock off the driver's & 0 & 3 & 1 & 4 \\
\quad the driver did not finish & 0 & 3 & 0 & 3 \\
\quad a relay delivered no line of its access log & 0 & 0 & 2 & 2 \\
\midrule
runs left & 58 & 277 & 51 & 386 \\
mean share of a run's actions in no log & 0.000 & 0.068 & 0.029 &  \\
mean share of a plane's actions missed by \texttt{spans} & 0.000 & 0.190 & 0.311 &  \\
mean share of a plane's actions missed by \texttt{access} & 0.000 & 0.300 & 0.273 &  \\
runs left in which the logs together hold every action & 58 & 2 & 18 &  \\
actions in a run, least to most & 316 to 353 & 316 to 353 & 157 to 302 &  \\
length of a run in s, least to most & 115 to 134 & 113 to 133 & 114 to 134 &  \\
\bottomrule
\end{tabular}
\par\smallskip{\footnotesize Conditions: \texttt{s0}, no stress; \texttt{s1}, both relays killed together and restarted together 15 to 45 s later; \texttt{s2}, each relay on its own outage and one originator killed for good. The integrity check is applied before any test. A run is counted once under each reason it has. The shares are over the runs left; the last two rows are computed for this table. Sources: \texttt{artifacts/robustness/study5-verdicts.json}, \texttt{artifacts/study5/confirm.json}.}

%% file: tables/study5_s1.tex
{\footnotesize\setlength{\tabcolsep}{3.5pt}\begin{tabular}{@{}llrrrrrrrrrr@{}}
\toprule
method & role & covered & Wilson 95\% & \shortstack[r]{un-\\bounded} & width & \shortstack[r]{abs.\\error} & error & \shortstack[r]{vs.\\\texttt{seq\_gap}\\\texttt{\_tail}} & \shortstack[r]{vs.\\\texttt{union}\\\texttt{\_count}} & \texttt{s0} & \texttt{s2} \\
\midrule
\texttt{ll\_sa\_hom} & primary & 269 & 0.944 to 0.985 & 0 & 32.0 & 3.9 & $-$0.5 & 223 & 261 (3) & 58 & 49 \\
\texttt{ll\_sa\_hom\_sig} & \begin{tabular}[t]{@{}l@{}}primary,\\no id trusted\end{tabular} & 269 & 0.944 to 0.985 & 0 & 32.0 & 3.9 & $-$0.5 & 223 & 261 (3) & 58 & 49 \\
\texttt{ll\_bic\_hom} & beside it & 262 & 0.913 to 0.967 & 2 & 32.0 & 4.2 & $-$0.1 & 217 & 263 & 58 & 49 \\
\texttt{ll\_average} & beside it & 266 & 0.930 to 0.978 & 38 & 43.0 & 4.0 & $+$1.0 & 188 & 223 (9) & 58 & 50 \\
\texttt{strat\_bic} & beside it & 255 & 0.883 to 0.947 & 10 & 26.0 & 4.0 & $+$0.6 & 208 & 244 (8) & 58 & 40 \\
\texttt{chapman\_te} & beside it & 250 & 0.862 to 0.932 & 0 & 23.9 & 4.0 & $-$0.7 & 224 & 261 (3) & 58 & 48 \\
\texttt{ll\_bic\_boot} & beside it & 243 & 0.833 to 0.911 & 2 & 26.6 & 4.2 & $-$0.1 & 217 & 263 & 58 & 42 \\
\texttt{ll\_bic} & beside it & 237 & 0.809 to 0.892 & 2 & 23.0 & 4.2 & $-$0.1 & 217 & 263 & 58 & 44 \\
\midrule
\texttt{chapman\_sa} & \begin{tabular}[t]{@{}l@{}}no end state\\(Q3)\end{tabular} & 10 & 0.020 to 0.065 & 0 & 3.6 & 50.7 & $-$54.3 & 4 (1) & 3 & 58 & 9 \\
\midrule
\texttt{seq\_gap\_tail} & baseline & 34 & 0.089 to 0.167 & 0 & 5.5 & 14.3 & $-$21.2 & -- & -- & 58 & 18 \\
\texttt{seq\_gap} & baseline & 0 & 0.000 to 0.014 & 0 & 0.0 & 18.6 & $-$24.4 & -- & -- & 58 & 15 \\
\midrule
\texttt{trust\_trace} & floor & 0 & 0.000 to 0.014 & 0 & 0.0 & 62.0 & $-$63.2 & -- & -- & 58 & 0 \\
\texttt{union\_count} & floor & 2 & 0.002 to 0.026 & 0 & 0.0 & 22.0 & $-$22.6 & -- & -- & 58 & 18 \\
\texttt{union\_bound} & floor & 277 & 0.986 to 1.000 & 277 & -- & 22.0 & $-$22.6 & -- & -- & 58 & 51 \\
\bottomrule
\end{tabular}}
\par\smallskip{\footnotesize The 277 runs left of 300 planned. Covered: runs whose 95\% interval holds the true count; an interval with no upper end (unbounded) counts unless its lower end is above the truth. Abs.\ error: median, in actions; error: mean, signed. Vs.: runs in which the point is strictly closer to the true count than that baseline's (ties in brackets, counted as losses). Only the primary estimator and \texttt{chapman\_sa} carry a criterion. Width: median width of the bounded intervals, computed for this table, not by the registered analysis. Last columns: runs covered of the 58 left in \texttt{s0} and 51 left in \texttt{s2}, neither with a criterion. Sources: \texttt{artifacts/robustness/study5-verdicts.json}, \texttt{artifacts/study5/confirm.json}.}

%% file: tables/study5_predictions.tex
\resizebox{\linewidth}{!}{\begin{tabular}{lp{6.4cm}p{8.4cm}l}
\toprule
 & registered prediction & result &  \\
\midrule
gate & \texttt{s1} is exercised: mean share of actions in no log 0.03 to 0.09 & 0.068; neither gate fires & right \\
P1, coverage & not met: Wilson lower end under 0.90 & the clause passes on 277 of the 300 planned runs, 23 set aside; ``not shown'' with those counted as not covered (P1 in the text) & \textbf{wrong} \\
P1, unbounded & share of unbounded intervals under 0.05 & 0 of 277 & right \\
P2, \texttt{seq\_gap\_tail} & pass & closer in 223 of 277; exact lower bound 0.753 & right \\
P2, \texttt{union\_count} & pass & closer in 261 of 277 with 3 ties; exact lower bound 0.908 & right \\
Q3 & ``missed'' on P1 for \texttt{chapman\_sa} & 10 of 277, Wilson 0.020 to 0.065; coverage reads ``missed'' & right \\
control & no interval above the truth in \texttt{s0} & 0 of 58 & right \\
\texttt{s2} & none made & described only & -- \\
\bottomrule
\end{tabular}}
\par\smallskip{\footnotesize Every result is on the runs left: 277 of the 300 planned \texttt{s1} runs (23 set aside on integrity) and 58 of the 60 planned \texttt{s0} runs. The registration marks the P1 prediction wrong if the coverage clause passes. Source: \texttt{artifacts/robustness/study5-verdicts.json}.}

%% file: tables/study5_set_aside.tex
{\footnotesize\setlength{\tabcolsep}{3.5pt}\begin{tabular}{@{}lrrrrrrrr@{}}
\toprule
 & \shortstack[r]{runs\\planned} & \shortstack[r]{set\\aside} & \shortstack[r]{of them\\\texttt{s1}} & late call & not ready & clock off & driver stopped & no access line \\
\midrule
\texttt{s0} & 60 & 2 & -- & 2 & 1 & 0 & 0 & 0 \\
\texttt{s1} & 300 & 23 & -- & 17 & 4 & 3 & 3 & 0 \\
\texttt{s2} & 60 & 9 & -- & 7 & 1 & 1 & 0 & 2 \\
all & 420 & 34 & 23 & 26 & 6 & 4 & 3 & 2 \\
\midrule
runs 1 to 307 & 307 & 7 & 4 & 6 & 0 & 1 & 0 & 1 \\
runs 308 to 322 & 15 & 4 & 4 & 3 & 0 & 2 & 0 & 0 \\
run 323 & 1 & 1 & 1 & 0 & 0 & 0 & 1 & 0 \\
runs 324 to 378 & 55 & 21 & 14 & 17 & 6 & 1 & 2 & 0 \\
runs 379 to 420 & 42 & 1 & 0 & 0 & 0 & 0 & 0 & 1 \\
\bottomrule
\end{tabular}}
\par\smallskip{\footnotesize Counts of runs. Upper block: by condition. Lower block: by place in the order the runs were made, which interleaves the three conditions, in the five stretches the decision log records; run 323 was under way when Docker Desktop quit. A run is counted once under each reason it has, so the reasons add to more than the runs set aside. Reasons: late call, a call made more than 0.25 s late; not ready, an originator not ready at the start; clock off, the containers' clock off the driver's; driver stopped, the driver did not finish; no access line, a relay delivered no line of its access log. No coverage, estimate or error is computed for any stretch. Sources: \texttt{runs\_set\_aside} in \texttt{artifacts/study5/confirm.json}, and the order \texttt{runs()} gives in \texttt{scripts/study5\_runs.py}, checked against \texttt{artifacts/study5/registered.json}.}

%% file: tables/study5_host.tex
\begin{tabular}{lrrrrrr}
\toprule
when a run started & runs & long & short & short, \texttt{s0} & short, \texttt{s1} & short, \texttt{s2} \\
\midrule
up to the last reading on battery (11\% charge) & 314 & 6 & 3 & 0 of 45 & 2 of 224 & 1 of 45 \\
from then until external power is plugged in & 58 & 1 & 17 & 2 of 9 & 11 of 41 & 4 of 8 \\
after external power is plugged in & 45 & 1 & 1 & 0 of 6 & 1 of 32 & 0 of 7 \\
\bottomrule
\end{tabular}
\par\smallskip{\footnotesize Post hoc. Counts of runs, by when a run started against the power log of the machine that made it; no estimate is read. Long: set aside with a call 20 s or more late, or a clock 20 s or more off (8 runs). Short: set aside for less than that (21 runs). Not in the runs column: the 3 runs that left no record (417 + 3 = 420 planned). In the runs column and in neither long nor short: the 2 runs whose relay delivered no line of an access log. Of the 8 long runs, 3 were made before the retained log begins and the other 5 are each beside a sleep of the machine; each of the 6 sleeps the log records while runs were made is beside one of them. Sources: \texttt{artifacts/study5/posthoc3.json}, \texttt{artifacts/study5/host/pmset-2026-11-16.txt}.}